\documentclass{article}

\usepackage[final]{corl_2019} 

\usepackage{subcaption}
\usepackage{graphicx}
\usepackage{amsmath}
\usepackage{amsfonts}
\usepackage{amsthm}
\usepackage{tabularx}
\usepackage{algorithmic}
\usepackage{algorithm2e}
\usepackage{multirow}

\newtheorem{proposition}{Proposition}

\title{Multi-Agent Reinforcement Learning with Multi-Step Generative Models}

\author{
  Orr Krupnik\\
  Technion \\
  \texttt{ok1@campus.technion.ac.il} \\
   \And
   Igor Mordatch \\
   Google Brain \\
  \texttt{imordatch@google.com} \\
   \And
   Aviv Tamar \\
   Technion \\
  \texttt{avivt@technion.ac.il} \\
}

\begin{document}
\maketitle

\begin{abstract}
We consider model-based reinforcement learning (MBRL) in 2-agent, high-fidelity continuous control problems -- an important domain for robots interacting with other agents in the same workspace. For non-trivial dynamical systems, MBRL typically suffers from accumulating errors. Several recent studies have addressed this problem by learning latent variable models for trajectory segments and optimizing over behavior in the latent space. In this work, we investigate whether this approach can be extended to 2-agent competitive and cooperative settings. The fundamental challenge is how to learn models that capture interactions between agents, yet are disentangled to allow for optimization of each agent behavior separately. We propose such models based on a disentangled variational auto-encoder, and demonstrate our approach on a simulated 2-robot manipulation task, where one robot can either help or distract the other. We show that our approach has better sample efficiency than a strong model-free RL baseline, and can learn both cooperative and adversarial behavior from the same data.

\end{abstract}



\section{Introduction}
	
Many real world problems involve multiple agents operating in the same environment, including autonomous vehicles \cite{lee2017desire,chen2017socially,schmerling2017multimodal}, navigation in presence of humans \cite{ivanovic2018generative}, learning communication \cite{mordatch2017emergence}, and multiplayer games \cite{stone2000layered,silver2016mastering}. In this work, we consider continuous control problems with two agents, a setting relevant when robots interact with other robots or with humans in their workspace.


When the system dynamics are unknown or hard to model, reinforcement learning (RL) can be used for learning control~\cite{sutton1998reinforcement}.
Research in RL can be classified into two main approaches: model-free RL (MFRL) and model-based RL (MBRL). In MFRL, agents learn a policy -- a mapping directly from states to actions, tuned to solve a specific task \cite{sutton1998reinforcement}. In MBRL, agents first learn a model of environment dynamics, and then plan optimal control using the learned model. An immediate advantage of MBRL is enabling control in tasks for which agents were not specifically trained. In addition, MBRL approaches are generally more sample-efficient and interpretable~\cite{deisenroth2011pilco,kurutach2018model}.

The typical approach for model learning involves predicting a single next state conditioned on past states and actions \cite{abbeel2005learning}. This method has some significant drawbacks, such as tending to accumulate error when rolled out multiple time steps into the future. For single-agent MBRL, several recent studies proposed to mitigate the error accumulation problem by learning deep generative models of multi-step trajectory segments~\cite{mishra2017prediction,ke2018modeling,hafner2018learning,rhinehart2018deep}. Here, we collectively refer to these methods as \emph{multi-step generative models}. The idea behind these approaches is to learn a distribution of future trajectory segments in an unsupervised manner, using a deep generative model (e.g. a variational auto-encoder, VAE \cite{kingma2013auto}), and perform planning by searching in the latent space of the model.
In essence, the latent space captures a compact `strategy space' of the agent, and it has been shown that optimization over this latent space is more effective than optimizing directly over actions~\cite{mishra2017prediction}.

When multiple agents are considered, game theory provides many guaranteed ways to solve multi-agent problems~\cite{cesa2006prediction}, but has mostly been studied in a discrete strategy space formulation, where single actions are played against other agents, e.g. in matrix games. The aim of our work is to combine such strategic decision making with high-fidelity problems that can be captured by RL formulations, such as continuous control domains.



In particular, we investigate whether multi-step generative models can be extended to 2-agent domains. Our main observation is that when optimizing over more than one agent, the generative model needs to admit two properties: it has to capture the interaction between the agents in the trajectory; and the latent space needs to be disentangled, such that the behavior of each agent can be optimized separately (as each agent may have a different objective). We discuss how to learn such models in various cases, and introduce a general algorithm for learning disentangled models using a variational lower bound on the mutual information. Using this idea, we develop a MBRL approach for either cooperative or competitive settings, based on Temporal Segment Models (TSMs)~\cite{mishra2017prediction}.

We demonstrate our approach in simulation on a continuous predator-prey domain, and on a 2-robot manipulation task. We show that using multi-step models overcomes the accumulating error problem of single-step MBRL, that our approach is more sample efficient than MFRL, and that we can use the same data to learn both cooperative and adversarial behavior. Thus, when implemented appropriately, multi-step latent variable models are a viable approach to multi-agent MBRL.

\section{Related Work}
Many approaches consider MBRL in domains with complex, stochastic dynamics. Our work is inspired by the Temporal Segment Models (TSMs) developed by~\citet{mishra2017prediction} and more recent approaches~\cite{ke2018modeling,hafner2018learning} which use conditional VAE (CVAE) based dynamics models to predict multiple time steps into the future, conditioned on a previous segment of states and actions. The dynamics model is then used to optimize over agent trajectories for a specific task. A different approach is taken by  \citet{nagabandi2017neural}, who use neural network dynamics models, combined with Model Predictive Control (MPC) and MFRL fine tuning to achieve effective control and stable gaits in the MuJoCo baseline environment. \citet{williams2017information} also use MPC to learn from a trained dynamics model, achieving high levels of performance on both simulated and actual hardware tasks. The models used, however, are single-time-step, fully connected neural network models.
Additionally, all the approaches described above consider only a single agent. 

Recently, MBRL approaches have also been successfully used to \textit{predict} the behavior of multiple agents in both cooperative and competitive environments. Most of these owe their success to advances in deep variational inference \cite{kingma2013auto}. \citet{lee2017desire} predict vehicle trajectories in environments with multiple counterparts, such as other vehicles and pedestrians. \citet{ivanovic2018generative} and \citet{schmerling2017multimodal} use CVAEs to predict human behavior for human-robot interaction. \citet{zheng2016generating} use deep RNN models to recreate trajectories of a team of basketball players. They provide the interesting observation of interpreting a latent variable as a `macro-goal', which controls the strategy of an agent for a given period of time. This is similar to our interpretation of the latent space as a `strategy space' for the agent to optimize in. To the best of our knowledge, our attempt to use deep generative models and the prediction of agent behavior to \emph{optimize} over trajectories for a given task in a multi-agent environment is, as of yet, a novel contribution. 


\section{Background}
A single-agent decision problem can be described as a Markov Decision Process (MDP, \cite{sutton1998reinforcement}). At each time step of an MDP, the system is in some state $x_t \in \mathcal{X}$ and agents can perform actions $u_t \in \mathcal{U}$, where $\mathcal{X}$ and $\mathcal{U}$ are the sets of all possible states and actions, respectively. The dynamics of the system are described by some function $f$ such that $x_{t+1} = f(x_t, u_t)$, or by some probability distribution $\mathcal{P}(x_{t+1} | x_t, u_t)$, governing the transitions from one state of the system to the next. Given a reward function $r(x)$, the typical goal is to maximize the expected sum of rewards $\mathbb{E}\left[\sum_{t=0}^T r(x_t)\right]$.

The Markov game~\cite{littman2001value} extends the MDP to incorporate multiple agents. For simplicity, we present the case of two agents, but our results can easily be extended to more players (e.g. as described by \citet{littman2001value}). Consider two agents $x$ and $y$ acting in a dynamic environment. At time-step $t$, the states of the agents are denoted by $x_t$ and $y_t$, respectively. The agents each select an action $u_t \in \mathcal{U}$ and $w_t \in \mathcal{W}$, out of their respective action sets. The dynamics of the environment are then governed by the distribution (or function) $\mathcal{P}(x_{t+1}, y_{t+1} \vert x_t, y_t, u_t, w_t)$, and the agents obtain rewards $r_x(x_t,y_t)$ and $r_y(x_t,y_t)$, respectively. This formulation can be used to define competitive scenarios (e.g. by selecting zero sum rewards), cooperative scenarios (e.g. by selecting the same reward function for both agents), or various combinations thereof. To simplify our presentation, in this work we focus on the case of deterministic dynamics, given by $x_{t+1}, y_{t+1} = f(x_t, y_t, u_t, w_t)$.\footnote{An extension to stochastic dynamics can easily be attained; see explanation following Proposition \ref{prop:saddle_point}.}



\subsection{Multi-step Generative Models}

The traditional approach for modelling dynamics attempts to approximate the transition function $f$~\cite{hunt1992neural,depeweg2017learning}. Multi-step generative models \citep{mishra2017prediction,rhinehart2018deep,ke2018modeling} learn a more generalized form of the system dynamics. Consider segments (or "chunks") of $H$ past agent observations and actions $X^-,U^-=\{x_{t-H+1},\dots,x_{t}\},\{u_{t-H+1},\dots,u_{t}\}$, and segments of future observations and actions $X^+,U^+=\{x_{t+1},\dots,x_{t+H}\},\{u_{t+1},\dots,u_{t+H}\}$. Multi-step models learn a distribution $P(X^+,U^+|X^-,U^-)$ to predict the future segment based on the past one, thus enabling consideration of delayed effects of previous actions and capturing longer temporal relations between states than a single-step model. 

In this work we focus on the TSMs of \citet{mishra2017prediction}, but the ideas presented here can be generalized to other latent variable models.
TSMs use a CVAE architecture~\cite{kingma2013auto}, where an encoder learns a distribution over a latent variable $Z$ conditioned on future observations and actions: $\mathcal{Q}(Z|X^+,U^+)$, 
while a decoder reconstructs $X^+,U^+$: 
$\hat X^+, \hat U^+ = D(X^-,U^-,Z)$.
The optimization objective is a variational lower bound of the data log likelihood~\cite{kingma2013auto}:
\begin{equation}
    \mathcal{L} = \mathbb{E}_{Z\sim Q(Z|X^+,U^+)} \left[\log P(X^+,U^+|X^-,U^-,Z)\right] -  \mathcal{D}_{KL}\left[Q(Z|X^+,U^+)||P_{prior}(Z)\right],
    \label{eq:elbo}
\end{equation}
where $P_{prior}(Z) = \mathcal{N}(0,I)$, and $P(X^+,U^+|X^-,U^-,Z) = \mathcal{N}(D(X^-,U^-,Z),I)$.
Each instance of the latent variable describes a segment of observations and actions, and the decoder can be used as a generative stochastic dynamics model, to create future segments using samples from the prior. 


One of the motivations for learning a dynamics model is to later use this model to solve the MDP. Assume that the total horizon of interest $T$ can be segmented to $k$ segments of length $H$. Then, the learned model for future segment distribution can be sampled $k$ times to produce a distribution for a $T$-length trajectory, which we denote as $\mathcal{P}^{(x)} (x_{t},u_t \vert x_{0:t-1},u_{0:t-1}, z^{(1:k)})$, where $z^{(1:k)}$ denotes $k$ samples of the latent variable $Z$ corresponding to the $k$ segments (i.e., each segment depends on one instance of the latent variable and on the previous segment, such that a probability of the full trajectory given the latent variables is well defined).
\citet{mishra2017prediction} suggest the following trajectory optimization problem:\footnote{In \citet{mishra2017prediction}, the action and state sequences were explicit, $X$ and $U$, and prediction consisted of two CVAEs: $P^{(u)}(U^+|U^-,Z)$ and $P^{(x)}(X^+|X^-, U^+, U^-,Z)$. Here, we unify the models into one CVAE. 
} 
\begin{equation}
    \max_{z^{(1:k)}} \mathbb{E} \left[ \sum_{t=1}^{T} r(x_{t}) \right],\quad 
    \textrm{s.t. } x_{t}, u_{t} \sim \mathcal{P}^{(x)} (x_{t},u_{t} \vert x_{0:t-1},u_{0:t-1}, z^{(1:k)}).
    \label{eq:traj_opt_clean}
\end{equation}
Note that the optimization is over the latent space $Z$ of the agent trajectory. In this sense, we are optimizing over the possible strategies that the agent can perform. 
Problem \eqref{eq:traj_opt_clean} can be solved using gradient descent.\footnote{The distribution $P(X^+|X^-,Z)$ is typically represented as a neural network that encodes the mean of a Gaussian, and the expectation in \eqref{eq:traj_opt_clean} is approximated with the deterministic prediction of the mean trajectory.} 



\section{Multi-Agent Temporal Segment Models}\label{sec:matsm}

Our goal in this work is to extend the TSM (and generally, multi-step generative models) to the Markov game setting. 
To simplify notation, we restrict our presentation to optimizing a single segment. The extension to a full trajectory is done by rolling out several segments, as in Eq.~\ref{eq:traj_opt_clean}. To further reduce clutter, we drop the conditioning on past segments in our notation.


Similarly to the definitions of $X^+,U^+$, let $Y^+,W^+$ denote the segments of $H$ future observations and actions for agent $y$. For a two player game, we would like to learn a generative model for the distribution $P(X^+,U^+,Y^+,W^+)$ with two latent components, $Z_x$ and $Z_y$, which correspond to the sequences of actions in a segment for agent $x$, $U^+$, and agent $y$, $W^+$, 
respectively.
Such a representation would allow us to cast the trajectory optimization problem in a standard game theoretic formulation, where the spaces of $Z_x$ and $Z_y$ denote the possible strategies for each agent. Thus, for a competitive scenario we have the following optimization problem:

\begin{align}
    \begin{split}
    \max_{Z_x} \min_{Z_y} \mathbb{E}\left[ \sum_{t=1}^{H} r(x_{t},y_{t}) \right], \quad
    \textrm{s.t. } x_{t},u_t, y_{t},w_t \sim P(X^+,U^+,Y^+,W^+|Z_x, Z_y).
    \end{split}
    \label{eq:traj_opt_multi}
\end{align}

Similarly, for the cooperative setting, the $\min_{Z_y}$ in Eq.~\ref{eq:traj_opt_multi} would be replaced with $\max_{Z_y}$. Note that the competitive setting \emph{requires} us to have a disentangled latent representation for the behaviors of the two agents, since we need to maximize over $x$ and minimize over $y$. This is an important distinction from the single agent case, and in the following we explore latent variable models that can accommodate such structure.

\subsection{Conditional Multi-Step Generative Models} \label{sec:cond_msgm}
A straightforward disentangled model representation can be obtained by conditioning. Since $P(X^+,U^+,Y^+,W^+)=P(X^+,U^+)P(Y^+,W^+|X^+,U^+)$, we can learn two separate models: one for $P(X^+,U^+)$ with a latent variable $Z_x$, and one for $P(Y^+,W^+|X^+,U^+)$ with a latent variable $Z_y$. We term this a \emph{conditional} model, and note that, as desired, the latent variables $Z_x$ and $Z_y$ represent the strategies of $x$ and $y$ respectively, by definition. 

While the conditional model is intuitive, our first insight in this paper is that it is not necessarily compatible with the competitive optimization objective \eqref{eq:traj_opt_multi}. To see this, note that since we are optimizing over $Z_x$, and $P(X^+,U^+)$ does not depend on $Z_y$, we are effectively letting agent $x$ `play first' and choose its trajectory, while agent $y$ can only react to the trajectory of $x$ that has already been chosen.\footnote{Conversely, modelling the distribution as $P(Y^+,W^+)P(X^+,U^+|Y^+,W^+)$ advantages player $y$.} This advantage to player $x$ is demonstrated in a simple blocking example in Fig.~\ref{fig:cond_illustration}, and verified in our experiments (see Fig.~\ref{fig:bumpers} in the appendix). 

\begin{figure}
    \centering
    \includegraphics[width=0.6\textwidth]{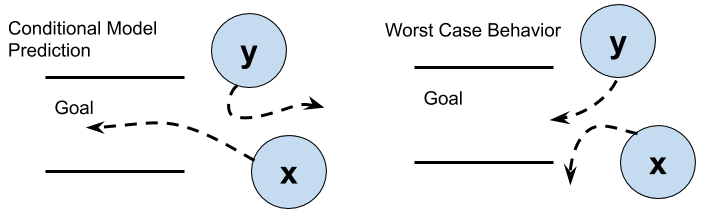}
    \caption{Optimistic predictions of the conditional model. In this illustrative example, the agents cannot pass through each other. Agent $x$'s goal is to enter the tunnel, and in the conditional model (left), it will always choose a trajectory that enters the tunnel, while $y$ would not be able to block it as it chooses a trajectory conditioned on $x$ having already entered the tunnel. In practice (right), $y$ can block $x$ from entering at all.}
    \label{fig:cond_illustration}
\end{figure}

We note, however, that the conditional representation can still be appropriate under some conditions. Consider a system where the dynamics of the agents are independent, i.e., can be written as $x_{t+1} = f_x(x_t, u_t)$ and $y_{t+1} = f_y(y_t, w_t)$, while the coupling between the agents in the task is realized only through the reward functions $r_x(x_t,y_t)$ and $r_y(x_t,y_t)$. In such cases, learning two independent models for the agents is compatible with the competitive optimization. 


\subsection{Disentangled Multi-Step Generative Models} \label{sec:disentangled}
The asymmetry issue of the conditional model motivates to study alternative disentangled models for the joint distribution $P(X^+,U^+,Y^+,W^+)$. We start by formally defining the disentanglement we desire in an idealized setting, and then propose two practical models to approximate this objective.

Consider the following max-min game, which represents the essence of our problem:
\begin{equation}\label{eq:prob_uw}
    \begin{split}
        \max_{U} \min_{W} \quad C(X,Y), \quad 
        s.t. \quad X,Y = F(U,W).
    \end{split}
\end{equation}
We view the multi-step generative model as some model for the action sequence with a parameters $z_x,z_y$, that is, we can write $U,W = G(z_x,z_y)$ for some $G$. Since the dynamics are assumed deterministic, we can also write $X,Y = (F \circ G)(z_x,z_y)$, and the objective is equivalent to $C \circ F \circ G$.

The next proposition shows a sufficient condition for optimization over the latent space as a substitute for problem \eqref{eq:prob_uw}. The proof, based on a simple chain rule, is given in the appendix, along with an extension to stochastic dynamics.

\begin{proposition}\label{prop:saddle_point}
Assume $G$ satisfies $\frac{\partial U}{\partial z_y} = 0$ and $\frac{\partial W}{\partial z_x}=0$ for all $z_x,z_y$, and consider a saddle point $z_x^*,z_y^*$ of $C \circ F \circ G$. If  $\frac{\partial U}{\partial z_x} \neq 0$ and $\frac{\partial W}{\partial z_y} \neq 0$, then $u^*,w^*=G(z_x^*,z_y^*)$ is a saddle point of $C \circ F$. 
\end{proposition}

Proposition \ref{prop:saddle_point} conveys an intuitive idea: if the latent variable for $x$ is independent of the actions of $y$, and vice versa, then we can safely optimize over the latent space in a max-min sense. Of course, this result is for an idealized setting, where the mapping from latent space to actions is bijective, while the whole point of the latent space is to represent only the most likely actions as seen in the data. Still, Proposition \ref{prop:saddle_point} provides a form of disentanglement suitable for the max-min game. We next ask \emph{how to learn} deep generative models that satisfy such disentanglement, and propose two approximations.

\subsubsection*{Disentanglement via Gradient Penalty}
A simple approach to training a disentangled model based on Proposition \ref{prop:saddle_point} is to add to the generative training loss a term of the form:
\begin{equation}\label{eq:gradient_loss}
    L_{\text{gradient}} = \| \nabla_{z_y} U \| + \| \nabla_{z_x} W \|.
\end{equation}
The loss in \eqref{eq:gradient_loss} directly corresponds to Proposition \ref{prop:saddle_point}. In practice, however, backpropagation through the gradient leads to a slow training procedure. We found that a proxy based on mutual information works well in practice and is much faster to train.

\subsubsection*{Disentanglement via Mutual Information}
Recent studies in learning disentangled latent spaces for image generation proposed to induce disentanglement by maximizing mutual information between latent and observation~\cite{chen2016infogan,klys2018learning}. We borrow these ideas, and propose to use the notion of mutual information to induce dependency between elements in the latent space and actions in the predicted trajectory. 

The mutual information between random variables $Q$ and $V$ is:
$
    I(Q;V) = H(Q) - H(Q|V) = H(V) - H(V|Q),
$
where $H(Q)$ denotes the entropy of $Q$, and $H(Q|V)$ denotes the conditional entropy~\cite{cover2012elements}. Following the discussion above, we would like $U$ to be independent of $Z_y$, and $W$ to be independent of $Z_x$. We propose to do this by \emph{maximizing} the mutual information between $Z_x$ and $U$, and between $Z_y$ and $W$. The intuition is that if $Z_x$ is maximally informative about $U$, then $U$ has to be independent of $Z_y$, since any dependence on $Z_y$ cannot be explained by $Z_x$, and vice versa.
We therefore propose the following loss term:
\begin{equation}\label{eq:loss_info}
    L_{\text{info}} = - I(z_x; U) - I(z_y; W)
\end{equation}
To optimize \eqref{eq:loss_info} in practice, we use the lower bound proposed by \citet{chen2016infogan}. We describe the full algorithm and architecture in the appendix. We observe that this disentangled model strikes a balance between ease of training and performance, and verify that this model solves a problem in which the conditional model fails (see appendix).

\section{Experiments}
We now present experiments in various environments, empirically showing that: 
\begin{enumerate}
    \item In the 2-agent setting, our multi-step approach improves over single-step dynamics models, and produces more accurate dynamics predictions, realizable in the environment.
    \item Our MBRL approach is more sample-efficient than MADDPG \cite{lowe2017multi}, a strong MFRL baseline, and outperforms it on our tested tasks. 
    \item Agents using our trained models perform well in various scenarios, including competitive and cooperative tasks, using the same data (collected while learning one specific task). 
\end{enumerate}
To test our models, we use simulators and data collected from two domains:
Multi-Agent Particle Environments~\cite{mordatch2017emergence} -- a standard, versatile and configurable benchmark for multi-agent RL, and a more complex, 2-robot simulation environment, built in Unity ML-Agents \cite{juliani2018unity}.


\begin{figure}
    \centering
    \centering
    \begin{subfigure}[b]{0.3\textwidth}
        \includegraphics[width=0.8\textwidth]{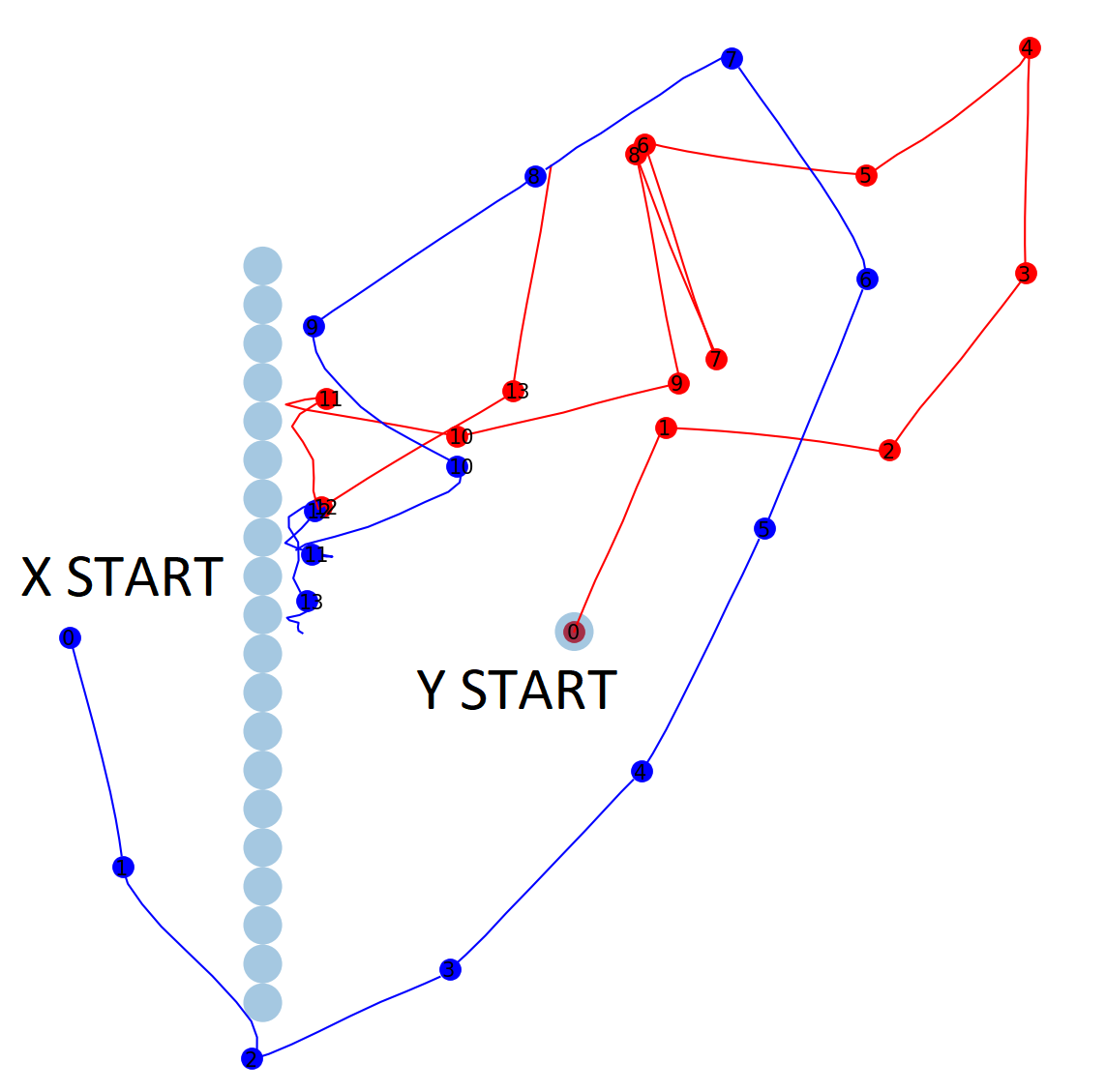}
        \caption{TSM}
        \label{fig:fc_vanilla}
    \end{subfigure} 
    \begin{subfigure}[b]{0.3\textwidth}
        \includegraphics[width=0.8\textwidth]{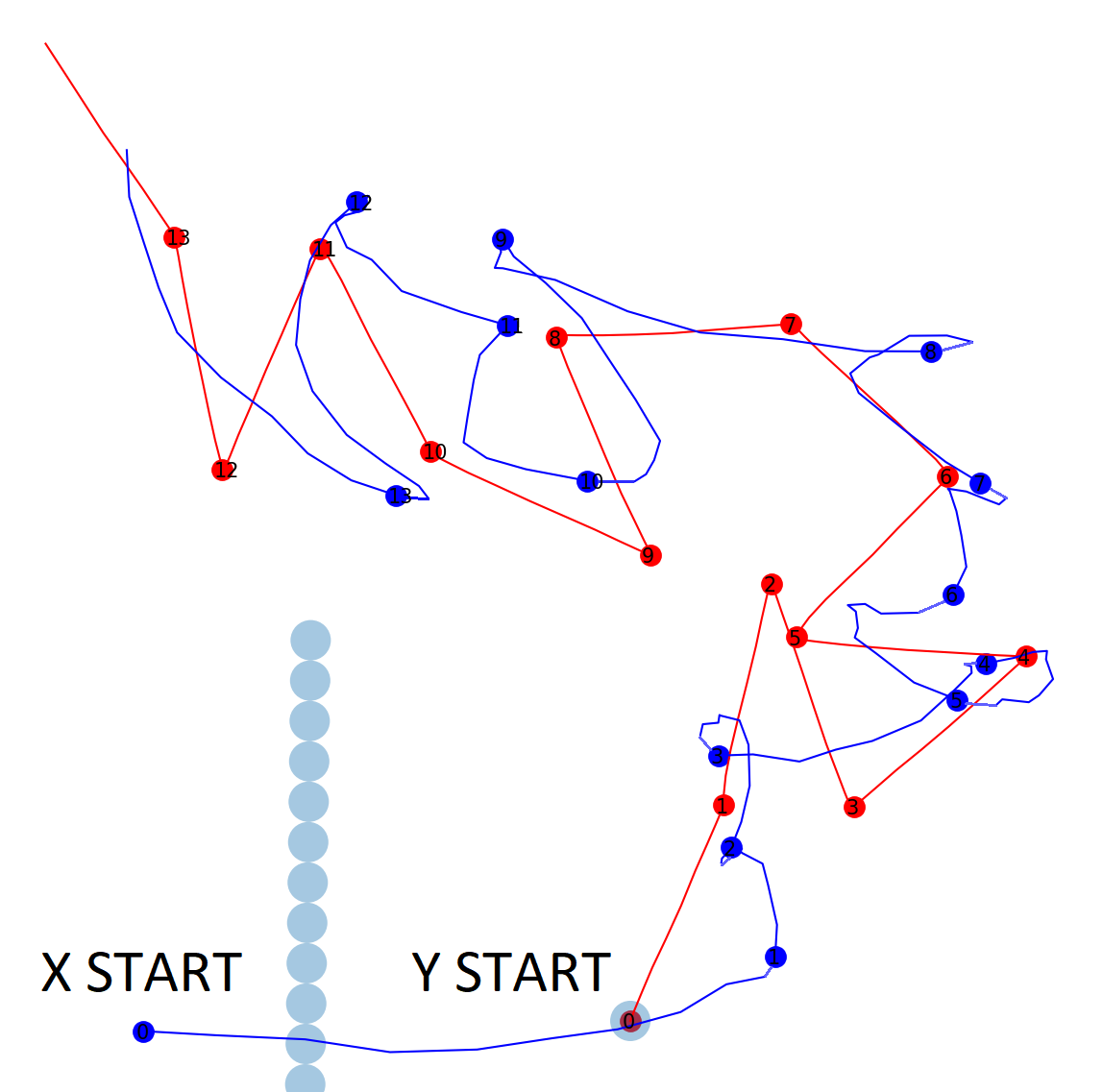}
        \caption{Single step baseline}
        \label{fig:ssb_traj}
    \end{subfigure}
    \begin{subfigure}[b]{0.3\textwidth}
        \includegraphics[width=\textwidth]{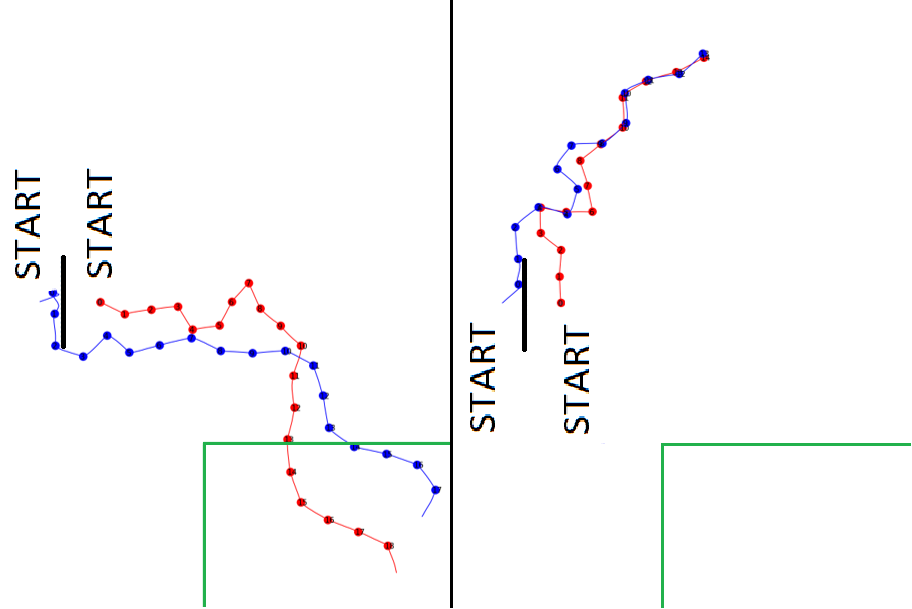}
        \caption{Safe-Zone domain}
        \label{fig:safe_zone}
    \end{subfigure}
    \caption{Predator-Prey domain. Predictions drawn from (a) our multi-step model, and (b) the single-step baseline. In both cases, the predator (blue) starts to the left of the wall obstacle, attempting to reach the prey (red), starting to the right. The trajectories are marked at every tenth time step to enable temporal comparison, coinciding with the segment lengths for our model. Both models \emph{predict} a successful chase of the prey. However, our agent successfully circumvents the obstacle,
    while the baseline exploits an error in the the single-step model and ignores the wall, which will fail at execution. (c) Predator-Prey domain, with a `safe zone' for prey. In this modified scenario, the prey has a `safe zone' (marked by the green box) where the predator gets no reward for capturing it. Left shows a prediction for the worst case prey behavior, while right shows a best case prediction.}  \label{fig:fc_traj_opt}
\end{figure}

\subsection{Comparison to a Single-step Model}\label{sec:single_step}


In the \textbf{predator-prey} environment, similar to the scenario created by \citet{lowe2017multi}, a predator agent chases an escaping target around an obstacle (see Fig.~\ref{fig:fc_traj_opt}). Predator reward is negative distance to the prey. In 
\textbf{`safe zone' predator-prey}, the prey agent has a `safe zone' at the bottom right corner of the game (see Fig.~\ref{fig:safe_zone}), where the distance rewards for the predator are nullified. 

We train a multi-step conditional model based on the TSM architecture \cite{mishra2017prediction} on data collected from a heuristic policy. We use our model to control the agent using an MPC-style approach~\cite{nagabandi2017neural}. Before executing each segment, we perform trajectory optimization \eqref{eq:traj_opt_multi} for the next 5 segments, and then execute the actions of the first segment of the optimal trajectory. For the other agent, we select the worst case (competitive) or best case (cooperative) segment predicted by our model (see Fig.~\ref{fig:safe_zone}).  
We compare our model to a single-step MLP dynamics model, using the same control method. 
Full details of our model architecture and training parameters, including the policy used for data collection, can be found in the appendix.

Fig.~\ref{fig:fc_vanilla} shows a sample from the vanilla predator-prey environment, where the predator starts to the left of the obstacle, chasing the prey starting on the right. Even in this relatively simple example, the shortfalls of the single-step model become apparent, as it exploits the model error to predict impossible trajectories that `cut through' the obstacle (see Fig.~\ref{fig:ssb_traj}). Table~\ref{tab:fc_opt_res} shows the cost incurred by the predator in the `safe zone' predator prey domain, which is the sum of distances to the prey, as long as the prey is not in the `safe zone'. Our model is more successful than the single-step baseline, due to the baseline exploiting model errors in its optimization, and then failing to execute its unrealistic predicted trajectories. For comparison, we present the reward collected by the semi-random exploration policy used for data collection (described in the appendix). 

These results confirm that the advantages of learning multi-step transition models, as established by \citet{mishra2017prediction} for the single agent case, also carry over to the multi-agent setting.

\begin{table}
    \caption{Results (average distance to prey, lower is better) for predator-prey domain. 
    Each row represents a different initial position of predator and prey relative to the wall obstacle. The results were obtained by averaging over 20 rollouts for each start position scenario. Our model outperforms the single-step baseline and the original (semi-random) exploration policy from both locations, with a larger advantage when the agents start on different sides of the wall (harder scenario). \\ }
    \centering
    \begin{tabular}{c||c|c|c}
     \textbf{Starting Positions (relative to wall)} & \textbf{Ours (conditional TSM)} & \textbf{Single Step} & \textbf{Random} \\
     \hline
     Same side & \textbf{0.7204} & 0.9223 & 1.5512 \\
     \hline
     Different sides & \textbf{0.9486} & 1.2537 & 1.7213
     \end{tabular}
     \label{tab:fc_opt_res}
\end{table}

\subsection{Comparison to a Model-Free Approach}\label{sec:mf_comp}

\begin{figure}[b]
    \begin{subfigure}[b]{0.49\textwidth}
        \includegraphics[width=0.95\textwidth]{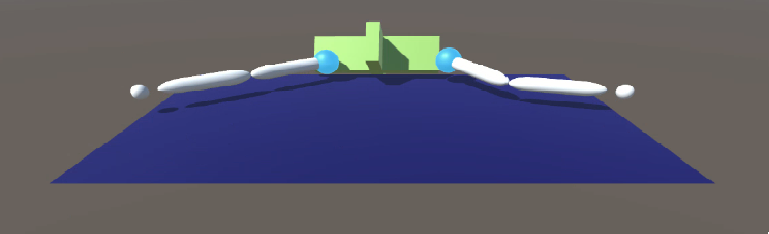}
        \caption{Cooperative: hitting together}
        \label{fig:real3}
    \end{subfigure} \hspace{0.01\textwidth}
    \begin{subfigure}[b]{0.49\textwidth}
        \includegraphics[width=0.95\textwidth]{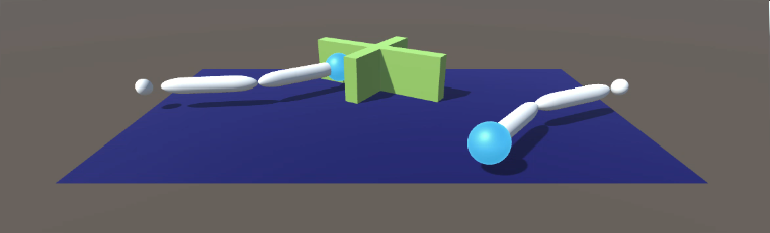}
        \caption{Cooperative: successive pushing}
        \label{fig:im3}
    \end{subfigure}
    \begin{subfigure}[b]{0.49\textwidth}
        \includegraphics[width=0.95\textwidth]{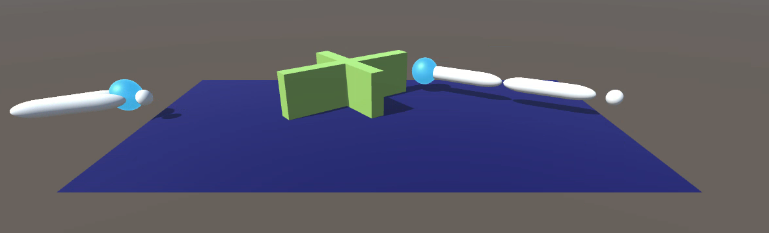}
        \caption{Competitive: object out of reach}
        \label{fig:real15}
    \end{subfigure} \hspace{0.01\textwidth}
    \begin{subfigure}[b]{0.49\textwidth}
        \includegraphics[width=0.95\textwidth]{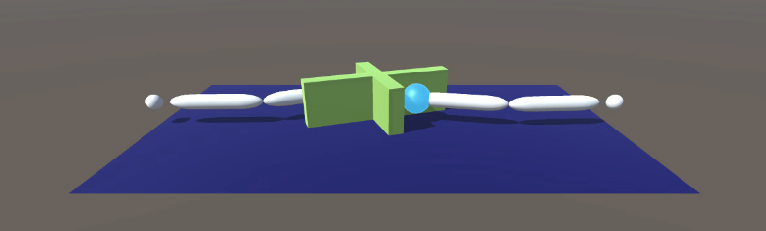}
        \caption{Competitive: pushing back}
        \label{fig:im15}
    \end{subfigure}
    \caption{Samples from the 2-robot Unity ML-Agents domain. (a) In the cooperative scenario, both agents (using our disentangled model) pushing the object together. (b) Same agents using a different strategy, of pushing the object successively. (c) In the competitive scenario, the left-hand MADDPG agent moves out of the way, as the object is out of reach of the right agent, thus the object remains on the platform indefinitely. (d) Left agent pushes back in the competitive scenario to prevent knocking the object off the platform. We encourage the reader to view the videos corresponding to these image samples, provided in the supplementary material.}
    \label{fig:reacher_samples}
\end{figure}

In both this subsection and the next, we use a simulated 2-robot manipulation task, where two manipulators interact with a movable object placed on a rectangular platform (see Fig.~\ref{fig:reacher_samples}). Each robot has two joints, which are each limited to movement around a single axis, such that the robot itself is limited to movement on a 2-D plain. This is an extension of the Reacher domain, as investigated in \cite{mishra2017prediction}, to the multi-agent case. We implement this domain in the Unity ML-Agents \cite{juliani2018unity} environment. We focus on two tasks: a cooperative game, where both agents try to push the object off the platform, and a competitive one described in the next section.

In the cooperative game, the observation space for each agent is a continuous feature vector describing the location, orientation and velocity of both agents and the manipulated object. As actions, each agent can apply continuous torques to both its joints. The reward is identical for both agents, and is comprised of a small negative reward for each time step, and a larger positive reward when the object is pushed off the platform, which also marks the end of a playing episode. 

Our goal is to compare our MBRL approach with MFRL. To provide a fair comparison, we use data for the MBRL method collected by running the MFRL algorithm. This guarantees that the results do not depend on the exploration strategy, but only on making the best use of the available data. As an MFRL algorithm, we chose MADDPG \cite{lowe2017multi}, a standard method for multi-agent environments. While training the MADDPG model, we collect the data accumulated in the replay buffer, and use pairs of past and future segments to train our disentangled multi-step model, described in Sec.~\ref{sec:disentangled}. 
Implementation details and parameters for both the MADDPG baseline and our model can be found in the appendix.

We used the mutual-information based disentangled model for all our results, due to its faster training time and suitability for both competitive and cooperative scenarios. Other disentangled models produced comparable results on the cooperative task, as reported in the appendix. We compare our model to the baseline at two points during its training, using the same collected data. The \textbf{short} model was trained from approximately 650k steps of MADDPG gameplay, and it is compared to the MADDPG model checkpoint saved at that time. The \textbf{long} model was trained using data from almost 3.3M time steps, and is compared to the appropriate, longer-trained baseline. As a MFRL approach, MADDPG supplies us with policies for both agents to use during test time. To control agents with our agents, we perform trajectory optimization \eqref{eq:traj_opt_multi} over the next 20 segments, and select the first segment of the optimal trajectory to play in the environment, proceeding using MPC. Table~\ref{tab:coop_results} shows the reward attained by each model. Our MBRL approach outperforms MADDPG, with a clearer advantage when trained with less data.

\begin{table}
    \caption{Reward for the cooperative 2-robot scenario, averaged over 20 episodes. Our model reaches better performance even with less training data, thus proving more sample-efficient. Best results are marked by bold text ($p < 0.05$ using Student's t-test). \\ }
    \centering
    \begin{tabular}{c||c|c}
     \textbf{Model} & \textbf{MFRL (MADDPG)} \cite{lowe2017multi} & \textbf{MBRL (ours, disentangled)} \\
     \hline
     \textbf{Short (650k steps)} & -1.79 $\pm$ 0.76 & \textbf{-0.72} $\pm$ \textbf{0.44} \\
     \hline
     \textbf{Long (3.3M steps)} & -0.79 $\pm$ 0.12 & -0.68 $\pm$ 0.42 
     \end{tabular}
     \label{tab:coop_results}
\end{table}

\subsection{Different Tasks Using the Same Training Data}
In the competitive scenario of the 2-robot environment, one agent tries to push the object off the platform while the other attempts to obstruct the object and keep it on the platform for as long as possible. Observation and action spaces for both agents are identical to the cooperative task, while the reward for the agent keeping the object on the platform is set to be adversarial, as the negative of the original reward described in the previous section. In essence, this scenario creates two distinct tasks: a \emph{pushing} task and a \emph{keeping} task, both with an adversary attempting to prevent success. 

We train the MADDPG baseline on this scenario with the same architecture and training configuration used for the cooperative game (while one of the agents receives the adversarial reward). Using the collected data, we train our disentangled multi-step model. Additionally, we use the multi-step model trained on data from the \emph{cooperative} task, optimizing for the adversarial reward. Results for the various combinations of agents can bee seen in Table~\ref{tab:comp_results}. Both our agents perform better than the MADDPG baseline in both tasks. Note that to use the model-free MADDPG agents in this scenario, we had to completely re-train them for a different task, while we can still use our model-based agent trained on data from the cooperative scenario to outperform the baseline. 

\begin{table}
    \caption{Reward for the competitive 2-robot scenario, averaged over 20 episodes. Left table: reward for the agent in the \emph{pushing} task against various adversaries. Right table: reward for the \emph{keeping} task. All MADDPG models were trained on the competitive scenario; the MBRL models are the mutual information disentangled version, where \textbf{co-op} denotes the `short' model described in Sec.~\ref{sec:mf_comp} (trained on data collected by MADDPG \emph{in the cooperative scenario}), and \textbf{comp} is the model trained on the same data as the baseline, collected in the competitive scenario. In both cases, our models score better than the MFRL baseline, even when using data collected for a different task. Best results in bold text ($p < 0.05$ using Student's t-test). }
    \centering
    \begin{tabular}{c|c|c||c|c|c}
     \multicolumn{3}{c||}{\textbf{"Push" Task}} & \multicolumn{3}{c}{\textbf{"Keep" Task}} \\
     \hline
     \textbf{Agent} & \textbf{Adversary} & \textbf{Score} & \textbf{Agent} & \textbf{Adversary} & \textbf{Score} \\
     \hline
     MADDPG & MADDPG & -6.43 $\pm$ 3.97 & MADDPG & MADDPG & 6.43 $\pm$ 3.97 \\
     MBRL (co-op) & MADDPG & -5.52 $\pm$ 3.41 & MBRL (co-op) & MADDPG & 7.77 $\pm$ 3.12 \\
     MBRL (comp) & MADDPG & \textbf{-3.03}$\pm$\textbf{3.54} & MBRL (comp) & MADDPG & 8.26 $\pm$ 2.81 \\
     
     \end{tabular}
     \label{tab:comp_results}
\end{table}

In addition to the quantitative results, Fig.~\ref{fig:reacher_samples} shows gameplay samples generated by our agents. The various images show different snapshots of strategy from both scenarios (cooperative and competitive), matching our observation that the latent variables encode some `strategy space' for the agents. 
\section{Conclusion}\label{sec:conclusion}
We extended the idea of multi-step generative models to handle competitive and cooperative 2-agent problems, showing its promise for complex, high-fidelity tasks relevant to real-world robotics environments. Key to our method is a novel disentangled generative model for 2-agent dynamics, and our results demonstrate improved performance and versatility over single-step and MFRL baselines.
In future work we intend to generalize these results to more than two agents, and apply these ideas to a real-world domain of human-robot interaction.
\acknowledgments{We thank the anonymous reviewers for their thoughtful comments and suggestions. This research was supported in part by a Google Faculty Research Award, and by a grant from the Open Philanthropy Project Fund, an advised fund of Silicon Valley Community Foundation.}
\clearpage
\bibliography{references}  

\begin{thebibliography}{29}
\providecommand{\natexlab}[1]{#1}
\providecommand{\url}[1]{\texttt{#1}}
\expandafter\ifx\csname urlstyle\endcsname\relax
  \providecommand{\doi}[1]{doi: #1}\else
  \providecommand{\doi}{doi: \begingroup \urlstyle{rm}\Url}\fi

\bibitem[Lee et~al.(2017)Lee, Choi, Vernaza, Choy, Torr, and
  Chandraker]{lee2017desire}
N.~Lee, W.~Choi, P.~Vernaza, C.~B. Choy, P.~H. Torr, and M.~Chandraker.
\newblock Desire: Distant future prediction in dynamic scenes with interacting
  agents.
\newblock In \emph{CVPR}, 2017.

\bibitem[Chen et~al.(2017)Chen, Everett, Liu, and How]{chen2017socially}
Y.~F. Chen, M.~Everett, M.~Liu, and J.~P. How.
\newblock Socially aware motion planning with deep reinforcement learning.
\newblock In \emph{IROS}, 2017.

\bibitem[Schmerling et~al.(2017)Schmerling, Leung, Vollprecht, and
  Pavone]{schmerling2017multimodal}
E.~Schmerling, K.~Leung, W.~Vollprecht, and M.~Pavone.
\newblock Multimodal probabilistic model-based planning for human-robot
  interaction.
\newblock \emph{arXiv preprint arXiv:1710.09483}, 2017.

\bibitem[Ivanovic et~al.(2018)Ivanovic, Schmerling, Leung, and
  Pavone]{ivanovic2018generative}
B.~Ivanovic, E.~Schmerling, K.~Leung, and M.~Pavone.
\newblock Generative modeling of multimodal multi-human behavior.
\newblock \emph{arXiv preprint arXiv:1803.02015}, 2018.

\bibitem[Mordatch and Abbeel(2018)]{mordatch2017emergence}
I.~Mordatch and P.~Abbeel.
\newblock Emergence of grounded compositional language in multi-agent
  populations.
\newblock In \emph{AAAI}, 2018.

\bibitem[Stone(2000)]{stone2000layered}
P.~Stone.
\newblock \emph{Layered learning in multiagent systems: A winning approach to
  robotic soccer}.
\newblock MIT Press, 2000.

\bibitem[Silver et~al.(2016)Silver, Huang, Maddison, Guez, Sifre, Van
  Den~Driessche, Schrittwieser, Antonoglou, Panneershelvam, Lanctot,
  et~al.]{silver2016mastering}
D.~Silver, A.~Huang, C.~J. Maddison, A.~Guez, L.~Sifre, G.~Van Den~Driessche,
  J.~Schrittwieser, I.~Antonoglou, V.~Panneershelvam, M.~Lanctot, et~al.
\newblock Mastering the game of go with deep neural networks and tree search.
\newblock \emph{nature}, 529\penalty0 (7587):\penalty0 484, 2016.

\bibitem[Sutton and Barto(1998)]{sutton1998reinforcement}
R.~S. Sutton and A.~G. Barto.
\newblock \emph{Reinforcement learning: An introduction}.
\newblock MIT press, 1998.

\bibitem[Deisenroth and Rasmussen(2011)]{deisenroth2011pilco}
M.~Deisenroth and C.~E. Rasmussen.
\newblock Pilco: A model-based and data-efficient approach to policy search.
\newblock In \emph{ICML}, 2011.

\bibitem[Kurutach et~al.(2018)Kurutach, Clavera, Duan, Tamar, and
  Abbeel]{kurutach2018model}
T.~Kurutach, I.~Clavera, Y.~Duan, A.~Tamar, and P.~Abbeel.
\newblock Model-ensemble trust-region policy optimization.
\newblock In \emph{ICLR}, 2018.

\bibitem[Abbeel and Ng(2005)]{abbeel2005learning}
P.~Abbeel and A.~Y. Ng.
\newblock Learning first-order markov models for control.
\newblock In \emph{NIPS}, 2005.

\bibitem[Mishra et~al.(2017)Mishra, Abbeel, and Mordatch]{mishra2017prediction}
N.~Mishra, P.~Abbeel, and I.~Mordatch.
\newblock Prediction and control with temporal segment models.
\newblock In \emph{ICML}, 2017.

\bibitem[Ke et~al.(2019)Ke, Singh, Touati, Goyal, Bengio, Parikh, and
  Batra]{ke2018modeling}
N.~R. Ke, A.~Singh, A.~Touati, A.~Goyal, Y.~Bengio, D.~Parikh, and D.~Batra.
\newblock Modeling the long term future in model-based reinforcement learning.
\newblock In \emph{International Conference on Learning Representations}, 2019.
\newblock URL \url{https://openreview.net/forum?id=SkgQBn0cF7}.

\bibitem[Hafner et~al.(2018)Hafner, Lillicrap, Fischer, Villegas, Ha, Lee, and
  Davidson]{hafner2018learning}
D.~Hafner, T.~Lillicrap, I.~Fischer, R.~Villegas, D.~Ha, H.~Lee, and
  J.~Davidson.
\newblock Learning latent dynamics for planning from pixels.
\newblock \emph{arXiv preprint arXiv:1811.04551}, 2018.

\bibitem[Rhinehart et~al.(2018)Rhinehart, McAllister, and
  Levine]{rhinehart2018deep}
N.~Rhinehart, R.~McAllister, and S.~Levine.
\newblock Deep imitative models for flexible inference, planning, and control.
\newblock \emph{arXiv preprint arXiv:1810.06544}, 2018.

\bibitem[Kingma and Welling(2014)]{kingma2013auto}
D.~P. Kingma and M.~Welling.
\newblock Auto-encoding variational bayes.
\newblock In \emph{ICLR}, 2014.

\bibitem[Cesa-Bianchi and Lugosi(2006)]{cesa2006prediction}
N.~Cesa-Bianchi and G.~Lugosi.
\newblock \emph{Prediction, learning, and games}.
\newblock Cambridge university press, 2006.

\bibitem[Nagabandi et~al.(2018)Nagabandi, Kahn, Fearing, and
  Levine]{nagabandi2017neural}
A.~Nagabandi, G.~Kahn, R.~S. Fearing, and S.~Levine.
\newblock Neural network dynamics for model-based deep reinforcement learning
  with model-free fine-tuning.
\newblock In \emph{ICRA}, 2018.

\bibitem[Williams et~al.(2017)Williams, Wagener, Goldfain, Drews, Rehg, Boots,
  and Theodorou]{williams2017information}
G.~Williams, N.~Wagener, B.~Goldfain, P.~Drews, J.~M. Rehg, B.~Boots, and E.~A.
  Theodorou.
\newblock Information theoretic {MPC} for model-based reinforcement learning.
\newblock In \emph{ICRA}, 2017.

\bibitem[Zheng et~al.(2016)Zheng, Yue, and Hobbs]{zheng2016generating}
S.~Zheng, Y.~Yue, and J.~Hobbs.
\newblock Generating long-term trajectories using deep hierarchical networks.
\newblock In \emph{NIPS}, 2016.

\bibitem[Littman(2001)]{littman2001value}
M.~L. Littman.
\newblock Value-function reinforcement learning in markov games.
\newblock \emph{Cognitive Systems Research}, 2\penalty0 (1):\penalty0 55--66,
  2001.

\bibitem[Hunt et~al.(1992)Hunt, Sbarbaro, {\.Z}bikowski, and
  Gawthrop]{hunt1992neural}
K.~J. Hunt, D.~Sbarbaro, R.~{\.Z}bikowski, and P.~J. Gawthrop.
\newblock Neural networks for control systems—a survey.
\newblock \emph{Automatica}, 28\penalty0 (6):\penalty0 1083--1112, 1992.

\bibitem[Depeweg et~al.(2017)Depeweg, Hern{\'a}ndez-Lobato, Doshi-Velez, and
  Udluft]{depeweg2017learning}
S.~Depeweg, J.~M. Hern{\'a}ndez-Lobato, F.~Doshi-Velez, and S.~Udluft.
\newblock Learning and policy search in stochastic dynamical systems with
  bayesian neural networks.
\newblock In \emph{ICLR}, 2017.

\bibitem[Chen et~al.(2016)Chen, Duan, Houthooft, Schulman, Sutskever, and
  Abbeel]{chen2016infogan}
X.~Chen, Y.~Duan, R.~Houthooft, J.~Schulman, I.~Sutskever, and P.~Abbeel.
\newblock Infogan: Interpretable representation learning by information
  maximizing generative adversarial nets.
\newblock In \emph{NIPS}, 2016.

\bibitem[Klys et~al.(2018)Klys, Snell, and Zemel]{klys2018learning}
J.~Klys, J.~Snell, and R.~Zemel.
\newblock Learning latent subspaces in variational autoencoders.
\newblock In \emph{Advances in Neural Information Processing Systems}, pages
  6445--6455, 2018.

\bibitem[Cover and Thomas(2012)]{cover2012elements}
T.~M. Cover and J.~A. Thomas.
\newblock \emph{Elements of information theory}.
\newblock John Wiley \& Sons, 2012.

\bibitem[Lowe et~al.(2017)Lowe, Wu, Tamar, Harb, Abbeel, and
  Mordatch]{lowe2017multi}
R.~Lowe, Y.~Wu, A.~Tamar, J.~Harb, P.~Abbeel, and I.~Mordatch.
\newblock Multi-agent actor-critic for mixed cooperative-competitive
  environments.
\newblock In \emph{NIPS}, 2017.

\bibitem[Juliani et~al.(2018)Juliani, Berges, Vckay, Gao, Henry, Mattar, and
  Lange]{juliani2018unity}
A.~Juliani, V.-P. Berges, E.~Vckay, Y.~Gao, H.~Henry, M.~Mattar, and D.~Lange.
\newblock Unity: A general platform for intelligent agents.
\newblock \emph{arXiv preprint arXiv:1809.02627}, 2018.
\newblock URL \url{https://github.com/Unity-Technologies/ml-agents}.

\bibitem[Van Den~Oord et~al.(2016)Van Den~Oord, Dieleman, Zen, Simonyan,
  Vinyals, Graves, Kalchbrenner, Senior, and Kavukcuoglu]{van2016wavenet}
A.~Van Den~Oord, S.~Dieleman, H.~Zen, K.~Simonyan, O.~Vinyals, A.~Graves,
  N.~Kalchbrenner, A.~Senior, and K.~Kavukcuoglu.
\newblock Wavenet: A generative model for raw audio.
\newblock \emph{arXiv preprint arXiv:1609.03499}, 2016.

\end{thebibliography}

\clearpage
\appendix
\section*{Appendix}
\subsection*{Proof of Proposition \ref{prop:saddle_point}}

\begin{proof}
Let $\ell = C\circ F \circ G$. 
A saddle point of $\ell$ satisfies $\frac{\partial \ell}{\partial z_x} = 0$ and $\frac{\partial \ell}{\partial z_y} = 0$, and (w.l.o.g.) $\frac{\partial^2 \ell}{\partial z_y^2} \succeq 0, \quad \frac{\partial^2 \ell}{\partial z_x^2} \preceq 0$. We have that
\begin{equation*}
    \frac{\partial \ell}{\partial z_y} = \frac{\partial C \circ F}{\partial U} \frac{\partial U}{\partial z_y} + \frac{\partial C \circ F}{\partial W} \frac{\partial W}{\partial z_y},
\end{equation*}
and therefore 
\begin{equation*}
    \frac{\partial C \circ F}{\partial W} \frac{\partial W}{\partial z_y} = 0.
\end{equation*}
So long as $z_y$ is not redundant, i.e., $\frac{\partial W}{\partial z_y}\neq 0$, we have that $\frac{\partial C \circ F}{\partial W}=0$. Following the same derivation for $z_x$, we obtain that if $\frac{\partial U}{\partial z_x}\neq 0$ then $\frac{\partial C \circ F}{\partial U}=0$.
We also have that 
\begin{equation*}
    \frac{\partial^2 \ell}{\partial z_y^2} = \frac{\partial^2 C \circ F}{\partial U^2}\left(\frac{\partial U}{\partial z_y}\right)^2 + \frac{\partial C \circ F}{\partial U}\frac{\partial^2 U}{\partial z_y^2} + \frac{\partial^2 C \circ F}{\partial W^2}\left(\frac{\partial W}{\partial z_y}\right)^2 + \frac{\partial C \circ F}{\partial W}\frac{\partial^2 W}{\partial z_y^2} \succeq 0,
\end{equation*}
and since $\frac{\partial U}{\partial z_y} = 0$ for all $z_y$, then $\frac{\partial^2 U}{\partial z_y^2}=0$, and we obtain (using our previous result $\frac{\partial C \circ F}{\partial W}=0$)
\begin{equation*}
    \frac{\partial^2 C \circ F}{\partial W^2}\left(\frac{\partial W}{\partial z_y}\right)^2 \succeq 0.
\end{equation*}
Since both matrices are symmetric and $\left(\frac{\partial W}{\partial z_y}\right)^2 \succeq 0$, we have that $\frac{\partial^2 C \circ F}{\partial W^2} \succeq 0$. Similarly, $\frac{\partial^2 C \circ F}{\partial U^2} \preceq 0$, and thus $u^*,w^*=G(z_x^*,z_y^*)$ is a saddle point for $C \circ F$.
\end{proof}

\subsection*{Extension to Stochastic Dynamics}
In the case of non-deterministic dynamics, we assume an independent variable $Q\sim P(Q)$ affects the dynamics $F$, such that $X, Y;Q=F(U,W;Q)$. The objective function may also depend on $Q$, and becomes $E_Q C(X,Y;Q)$. Note, however, that given the random $Q$, both the dynamics $F$ and the cost $C$ become deterministic functions of $X,Y$ and the random sample of $Q$. We can therefore denote the objective function $\ell_Q = E_Q [C\circ F \circ G]$. The proof then continues along the same lines, noting the fact that we can re-order expectation and differentiation in this case (since we assume continuous dynamics). Noting also that $G$ does not depend on $Q$, we have:
\begin{equation*}
    \frac{\partial \ell_Q}{\partial z_y} = E_Q \frac{\partial \ell}{\partial z_y} = \frac{\partial U}{\partial z_y} E_Q \frac{\partial C \circ F}{\partial U} + \frac{\partial W}{\partial z_y} E_Q \frac{\partial C \circ F}{\partial W},
\end{equation*}
and therefore 
\begin{equation*}
    \frac{\partial W}{\partial z_y} E_Q \frac{\partial C \circ F}{\partial W} = 0.
\end{equation*}
Following the same as above, we have that $E_Q \frac{\partial C \circ F}{\partial W}=0$ and $E_Q \frac{\partial C \circ F}{\partial U}=0$. Plugging into the second derivative, again re-ordering derivative and expectation, we obtain in the same manner:
\begin{equation*}
    \left(\frac{\partial W}{\partial z_y}\right)^2 E_Q \frac{\partial^2 C \circ F}{\partial W^2} \succeq 0.
\end{equation*}
Therefore, $E_Q \frac{\partial^2 C \circ F}{\partial W^2} \succeq 0$ and similarly $E_Q \frac{\partial^2 C \circ F}{\partial U^2} \preceq 0$, resulting in $u^*,w^*=G(z_x^*,z_y^*)$ being a saddle point for $E_Q [C\circ F]$, and our optimization in the latent space holding for the case of stochastic dynamics as well. 

Algorithmically, our CVAE architecture handles the stochastic dynamics by producing stochastic trajectories, the empirical average of which we can use to optimize our model. We also add a component to our latent space which is not limited by the mutual information or gradient losses, thus accounting for stochastic elements in the dynamics which can affect either one of the agents. A similar technique was used by \citet{chen2016infogan} to account for noise in image generation. 

\subsection*{Algorithm: Multi-Agent Model-Based Planning with Trajectory Optimization}
Below is the pseudo code for the algorithm described in Sec.~\ref{sec:disentangled}, explaining how to use our disentangled models for trajectory optimization. We use multiple restarts of the optimized latent variable as a way to batch the training and save some time -- instead of initializing a single instance of the random variable and optimizing over it for many iterations, we draw multiple instances, optimize over all of them in parallel for less iterations, and then select the best result. This algorithm describes the prediction and optimization of one segment; since we are using MPC \cite{nagabandi2017neural}, once we select a best segment we play it in the environment and repeat the selection process.  

Another thing to note is that this is the process for a single agent, considering the behavior of its counterpart in the environment. When playing in an actual simulator, the other agent can use a similar algorithm, reversing the roles of the optimized and sampled random variables. Additionally, this algorithm can easily be extended to more than two players by use of a disentangled model incorporating multiple additional agents. 

Note that the $N$ multiple restarts and $m$ target trajectories are a proxy for full $\max \min$ optimization -- instead of sampling one random variable for each agent and optimizing both the $\min$ and $\max$ parts in the respective latent spaces, we sample multiple instances, optimize less and select the best (or worst) outcome. This saves time during optimization, as it allows us to optimize over larger batches of variables, for less time steps. In exchange, however, it may reduce the performance of our method. 

\begin{algorithm}[H]
    \begin{algorithmic}[1]
        \STATE $x_0, y_0\leftarrow $ Initialize agents' states
        \STATE $N, m, l, k \leftarrow$ initialize batch size (number of optimization restarts), non-controllable agent trajectory samples, latent dimension of $Z$, segments per game
          \STATE $z_x \leftarrow$ initialize latent variable $(N \times k \times l )$ by sampling from $\mathcal{P}_{prior}(Z_x)$
          \STATE $z_y \leftarrow$ sample latent variable $(m \times k \times l )$ from $\mathcal{P}_{prior}(z_y)$
          \STATE opt $\leftarrow$ initialize  optimizer (Adam)
          \FORALL{restarts $i$ in $N$ (this part can be paralleled)}
            \FORALL{epochs}
                \FORALL{samples $j$ in $m$ (this loop can be paralleled)}
                    \FORALL{game segments}
                        \STATE $x^+_{i,j} = f^x(x^-_{i,j}, y^-_{i,j}; z_x^{(i)})$
                        \STATE $y^+_{i,j} = f^y(x^-_{i,j}, y^-_{i,j}, x^+_{i,j}; z_y^{(j)})$
                    \ENDFOR
                    \STATE loss$_{i,j} = -r_x(x_{i,j}$, $y_{i,j})$ summed over all segments
                \ENDFOR
            \ENDFOR
            \STATE Backprop loss
            \STATE opt.step()
        \ENDFOR
        \STATE Return trajectory $x_{i,j}$ that corresponds to the lowest $i$ in loss$_{i}$ and the highest $j$ in loss$_{i,j}$
    \end{algorithmic}
    \caption{Multi-Agent Trajectory Optimization}
    \label{alg1}
\end{algorithm}

Note that in line 21, we take the lowest $i$ in loss$_i$ and the highest $j$ in loss$_{i,j}$. This corresponds to the $\min \max$ game, suitable for an adversarial opponent. When playing a cooperative game, we take the lowest loss of both indices. 

\subsection*{Implementation Details for Conditional Predator-Prey Model}

The data we collected to train the world model consists of 30k trajectories of 50 steps each, where the agent applies 5 steps in a random direction, then continues to move with inertia, selects a new direction and repeats. This type of exploration provides trajectories which are relatively evenly distributed around the playing field, along with a sizable percentage of collisions with the obstacle, which are important for learning an accurate world model. We used an observation space reflecting the fact that agents rarely interact with each other in this relatively simple domain, and are dependent only in their reward. Each agent's observation space included its location, its velocity and a vector pointing to the middle of the obstacle. During training, we offset the agent location (but not the obstacle vector) by the first location in the segment. In this manner, we trained the same models for both predator and prey agents. 
Similarly to \citet{mishra2017prediction}, we used an encoder with two layers of dilated causal convolutions (32 channels each) and $\tanh$ activations, and a decoder with three dilated causal convolution layers (32 channels) and an additional 1-dimensional convolution at the output. The decoder activation functions are the same as suggested for this type of architecture \cite{van2016wavenet} and used by \citet{mishra2017prediction}:
\begin{align}
\tanh(W_{f,k} * s + V^T_{f,k}z) \odot \sigma(W_{g,k} * s + V^T_{g,k}z)
\end{align}
where $*$ denotes a convolution, $\odot$ is element-wise multiplication, $s$ is the output of the previous layer, $z$ is the latent variable and both $V$ and $W$ are learned weights. We found that a segment length of 10 steps and latent spaces of dimension 8 or 12 were a good choice for accurately learning the environment dynamics. We trained our model with Adam for 2000 epochs, using the standard hyper-parameters. We weighted the KL element of the VAE loss with $\beta=0.005$. 

As described in Sec.~\ref{sec:single_step}, we also trained a single-step baseline for this scenario. After conducting an architecture search, we concluded that a fully-connected, 3-layer MLP performed as well as any of the other architectures, and used it as a baseline for its simplicity of implementation. 

\subsection*{Implementation Details for MADDPG in the 2-Robot Environment}
We train MADDPG~\citep{lowe2017multi} using 3 layer MLPs for both the Actor and Critic networks, using 64 neurons per layer. We use Adam with the standard hyperparameters to optimize both the Actor and Critic networks, and train for 5000 episodes, each at most 2000 steps long (episodes can end earlier when the object is pushed off the platform). We use batches of size 512 to update the actor and critic networks every 100 steps. At each such update, we update the Actor and Critic networks 40 times, and then update the target networks once. Other implementation details and hyperparameters are similar to the ones in the original implementation, provided by \citet{lowe2017multi}.

\subsection*{Implementation Details for the Disentangled Model in the 2-Robot Environment}
In the 2-robot domain described in Sec.~\ref{sec:mf_comp}, the observation space for each agent is 21-dimensional, describing the rotation, position and velocity (both linear and angular) of its two parts, as well as the position of its hand and the hand of the second agent. The observation space also includes details about the position, rotation and velocity of the pushable object. 
Actions taken by the agents are torques applied to their joints, and are limited to $[-1,1]$. This gives each agent a 2-dimensional, continuous action space. Finally, we gave a reward of $-0.005$ for every time step, and a reward of $1.0$ when the object falls off the platform. For the adversarial agent in the competitive scenario, the reward is the negative of the regular one. 
 
The 2-robot environment is somewhat more complex than the multi-agent particle environments described above, and therefore we use a more expressive architecture for our models trained on this domain. For the encoder we kept the same architecture, but added another layer, and used wider layers -- 64 channels each. For the decoder, we used a similar architecture to the TSM \cite{mishra2017prediction} constructed of dilated causal convolutions. However, we used a deeper architecture, with residual blocks of 2 convolutions, each with 64 channels, in place of each single convolution in the original model. The decoder used three of these residual blocks, and an additional $1 \times 1$ convolutional layer at the end, with 128 channels. 
Additionally, to maximize mutual information between parts of the latent space and corresponding parts of the output (Sec.~\ref{sec:disentangled}), we added two additional encoder networks, encoding the output of the decoder back to match the latent space (as suggestetd by \citet{chen2016infogan}). These encoders have a similar convolutional architecture to the main encdoer, albeit with 16-channel layers. This adds a third, mutual information component to the CVAE loss term. 
 
For this task, we used 20-step segments, and a 16-dimensional latent space. We trained our models for 2000 epochs using Adam with a learning rate of $0.0001$. We weighted the KL term in the loss with $\beta=0.002$, and the mutual information term with $\gamma=0.001$. 
 
For trajectory optimization, we predicted 20 segments into the future, each 20 steps long. The maximum episode length when training MADDPG was 2000 steps, so this type of prediction covers $\frac{1}{5}$ of an episode. We used 10 multiple restarts for the latent variable ($N$ in Alg.~\ref{alg1}), 10 trajectories to select from for the opposing agent ($m$ in Alg.~\ref{alg1}) and optimized for 30 iterations before selecting the best segment. 

\subsection*{Additional Visualizations of Predicted Trajectories}

\begin{figure}[t]
    \centering
    \begin{subfigure}[b]{0.335\textwidth}
        \includegraphics[width=\textwidth]{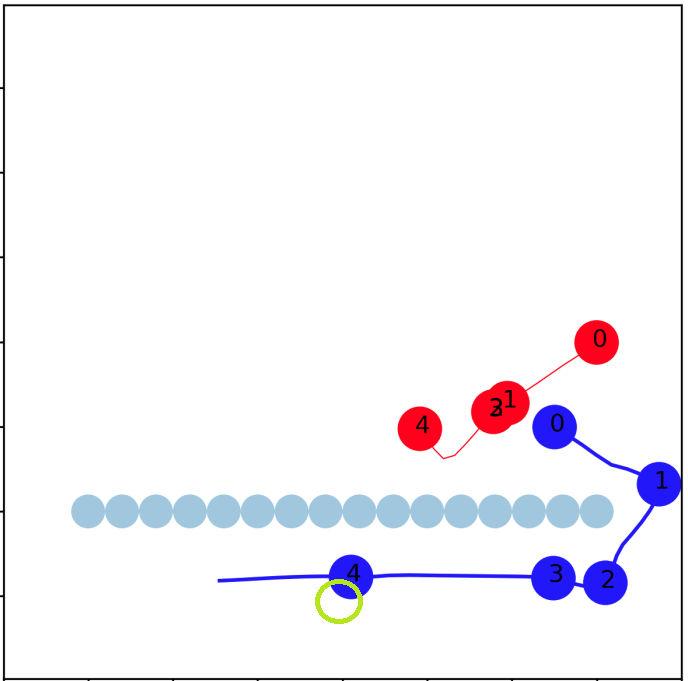}
        \caption{Conditional model}
        \label{fig:conditional}
    \end{subfigure} \hspace{0.05\textwidth}
    \begin{subfigure}[b]{0.335\textwidth}
        \includegraphics[width=\textwidth]{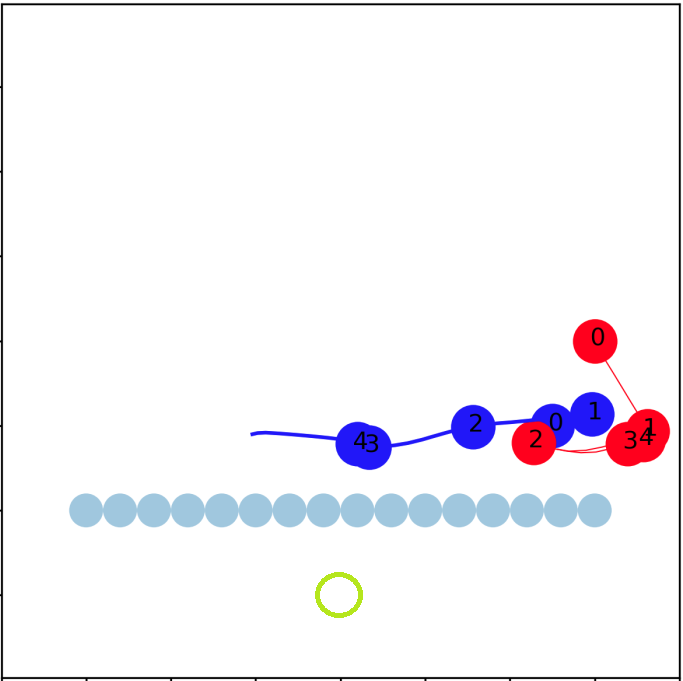}
        \caption{Disentangled model}
        \label{fig:disentangled}
    \end{subfigure}
    \caption{Comparison between disentangled and conditional models on box bumper domain. Blue agent ($x$) attempts to reach the target location below the horizontal obstacle (green circle), possibly being blocked by the red agent ($y$) in the narrow passage. Numbers indicate the segment order in time. Fig.~\ref{fig:conditional} shows a trajectory predicted by the conditional model. Since $x$ is independent of $y$, which only `responds' to its actions, the optimization cannot imagine a trajectory for which $x$ is blocked by $y$ en route to the goal, even in the worst case (risk-averse optimization). In Fig.~\ref{fig:disentangled}, a sample is shown from the predictions of the disentangled model, where $x$ chooses to go left, since it assumes $y$ will block it in the right passage.}
    \label{fig:bumpers}
\end{figure}

To test our hypothesis regarding the limits of the conditional model, we constructed a toy domain in the multi-agent particle environment \cite{lowe2017multi}, in which two agents, which are relatively large compared to the entire domain, move around a square playing area with a wall at the bottom part of it by applying forces in one of the four cardinal directions. Due to the limited size of this domain, the agents tend to collide often and block each other's way. In this domain, we used the data collection policy described for the predator-prey domain (see above in the appendix), modified such that the random direction selection is weighted towards the location of the other agent to promote collisions between the agents, which are an area of interest in the state space.
The task of one agent in this environment is to reach the goal location, which is located below the middle of the wall. The other agent may attempt to block its path. The observation space includes the locations and velocities of both agents.

The experiments in this domain are qualitative in nature, and serve as an example to the case where the conditional multi-step generative model (Sec.~\ref{sec:cond_msgm}) may not be suitable, and the disentangled models (Sec.~\ref{sec:disentangled}) may be necessary for a correct solution. We trained two models on this domain: the first was trained in the conditional form, where one controlled agent attempts to reach the goal, and the second agent plans trajectories conditioned on the path of the first; a second, disentangled model was trained using the mutual information maximization technique. Fig.~\ref{fig:bumpers} presents the results of trajectory prediction using both models. 

\begin{figure}
    \centering
    \begin{subfigure}[b]{0.4\textwidth}
        \includegraphics[width=\textwidth]{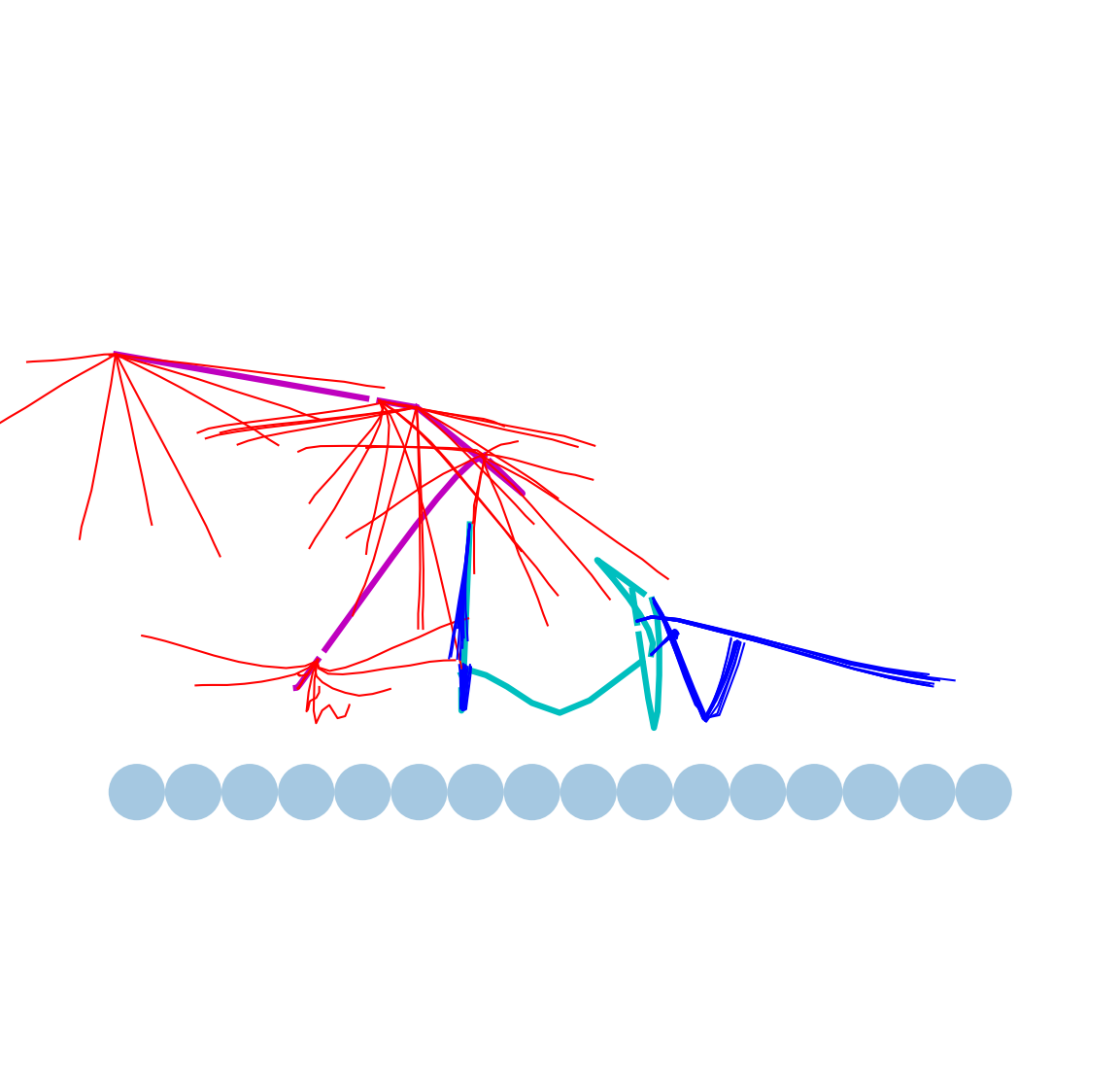}
        \label{fig:freeze1}
    \end{subfigure} \hspace{0.05\textwidth}
    \begin{subfigure}[b]{0.4\textwidth}
        \includegraphics[width=\textwidth]{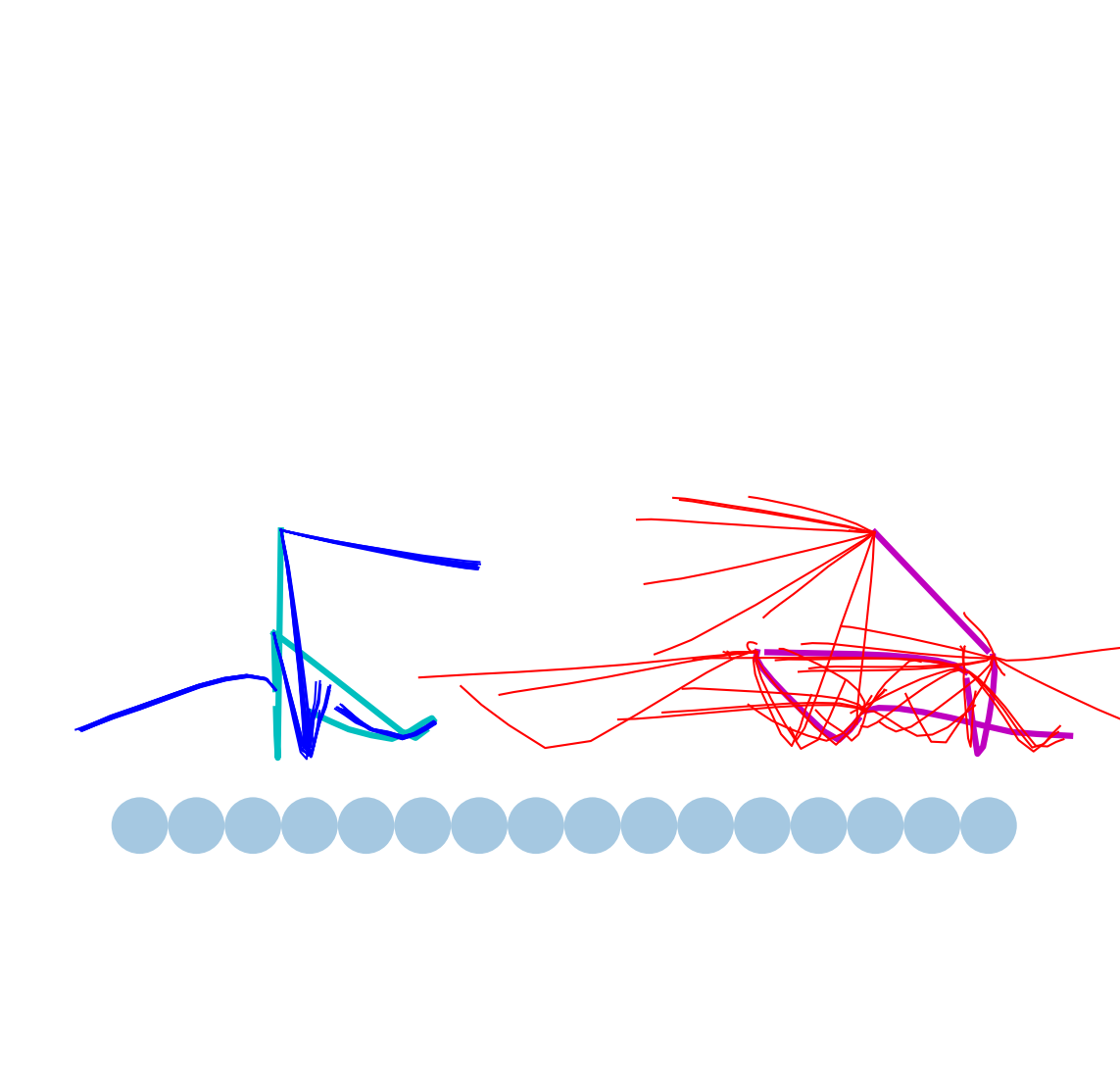}
        \label{fig:freeze2}
    \end{subfigure}
    \caption{Two examples sampled from the disentangled model with one part of the latent space `frozen' (set to a constant value), and the other part sampled from the prior. As expected, the behavior displayed is deterministic for one player, and distributed for the other. Red marks the distributed predictions for one agent, caused by sampling its latent variable from the prior. Blue trajectories are the ones predicted when the latent variable is 'frozen', thus causing deterministic behavior. Magenta and cyan lines denote the original trajectories collected from data, for reference. } \label{fig:freeze}
\end{figure}

Another experiment we conducted with this toy domain was designed to ensure that our disentangled model learns representations where different parts of the latent space control different parts of the output (i.e. different players). We run a simple experiment: we set one part of the latent space ($z_1$) to a constant value, and sample from the prior distribution to obtain values for the other part ($z_2$). We then use this latent variable construct, along with past segments, to predict future segments using our decoder. We expect to produce trajectories where one agent acts in a deterministic manner (since given $Z$, $D(X^-, Z)$ is a deterministic function) and the other agent producing a distribution over possible trajectories. Fig.~\ref{fig:freeze} shows example results of this experiment. 

\subsection*{Additional Results and Ablation Testing}
To complement the results of the cooperative 2-robot scenario presented in Sec.~\ref{sec:mf_comp}, we present here results obtained with other types of models, and with ablation testing for some of the training parameters. The cooperative results obtained in Sec.~\ref{sec:mf_comp} were obtained using 30 steps of optimization for each segment, 10 multiple restarts for the controlled agent and 10 samples for the other agent when conducting optimization. The 'no optimization' tests were run with 20 multiple restarts for the controlled agent, but no optimization at all (only selecting the best case trajectory). The `no multiple restarts' tests were run for 30 optimization steps, but with just one trajectory sampled. 

Additionally, we mentioned in Sec.~\ref{sec:disentangled} that other types of disentangled models, namely gradient-based and conditional models, performed similarly to the mutual-information based model. Table~\ref{tab:ablations} shows results for these other types of models, as well as the ablation tests described above. We repeat, for comparison, the results from Table~\ref{tab:coop_results} in the top section of this table.

\begin{table}
    \caption{Reward for the cooperative 2-robot scenario, averaged over 20 episodes, for various types of models with some ablations. \\ }
    \centering
    \begin{tabular}{l||c|c}
     \textbf{Model} & \textbf{Short (650k steps)} & \textbf{Long (3.3M steps)} \\
     \hline
     \hline
     \textbf{MADDPG (MFRL)} & -1.79 $\pm$ 0.76 & 0.79 $\pm$ 0.12 \\
     \textbf{Mutual Information} & -0.72 $\pm$ 0.44& \textbf{-0.68} $\pm$ \textbf{0.42}\\
     \textbf{Conditional} & -0.73 $\pm$ 0.44 & -1.35 $\pm$ 2.07 \\
     \textbf{Gradient Based} & -0.85 $\pm$ 0.33 & -1.11 $\pm$  0.55 \\
     \hline
     \textbf{Mutual Information (no optimization)} & \textbf{-0.71} $\pm$ \textbf{0.42} & -0.97 $\pm$ 0.89 \\
     \textbf{Conditional (no optimization)} & -0.83 $\pm$ 0.62 & -1.75 $\pm$ 2.79\\
     \textbf{Gradient Based (no optimization)} & -0.94 $\pm$ 0.44 & -1.06 $\pm$ 2.10 \\
     \hline
     \textbf{Mutual Information (no multiple restarts)} & -2.25 $\pm$ 3.29 & -1.10 $\pm$ 0.76 \\
     \textbf{Conditional (no multiple restarts)} & -1.24 $\pm$ 2.05 & -1.14 $\pm$ 2.17 \\
     \end{tabular}
     \label{tab:ablations}
\end{table}

As mentioned in Sec.~\ref{sec:disentangled}, we elected to use the mutual-information-based model for its ease of training and suitability for the competitive model, even though results for the other models were comparable. It is worth noting that the multiple restart scheme produces results on par with the longer optimization setting, which may mean variety in the latent space could produce better strategies leading to better performance. We leave further analysis of the latent space to future work. 

\end{document}